\documentclass[10pt,twocolumn,letterpaper]{article}

\usepackage{cvpr}              
\usepackage[accsupp]{axessibility}  

\usepackage[table, dvipsnames]{xcolor}
\usepackage{multirow}
\usepackage{amsmath}

\usepackage{color}
\definecolor{crimson}{rgb}{0.86, 0.08, 0.24}
\definecolor{gray}{rgb}{0.5,0.5,0.5}
\definecolor{green}{rgb}{0, 0.4, 0}
\definecolor{mahogany}{rgb}{0.75, 0.25, 0.0}
\definecolor{purple}{rgb}{0.6, 0, 0.6}
\definecolor{darkgreen}{rgb}{0, 0.4, 0}
\definecolor{frenchblue}{rgb}{0.0, 0.45, 0.73}
\definecolor{magenta}{rgb}{1,0,1}
\definecolor{pink}{rgb}{1,0.412,0.706}
\definecolor{forestgreen}{RGB}{59, 192, 0}
\definecolor{goldenyellow}{RGB}{255, 192, 0}

\definecolor{lightyellow}{rgb}{1,1, 0.6}
\definecolor{lightorange}{rgb}{1, 0.8, 0.6}
\definecolor{lightred}{rgb}{1, 0.6, 0.6}

\newcommand{\ylwcell}[0]{\cellcolor{lightyellow}}

\long\def\ignorethis#1{}

\newcommand {\fong}[1]{#1}

\usepackage{indentfirst}
\usepackage{fixmath}

\definecolor{green}{rgb}{0.0, 0.5, 0.0}
\definecolor{blue}{rgb}{0.0, 0.0, 0.9}

\definecolor{cvprblue}{rgb}{0.21,0.49,0.74}
\usepackage[pagebackref,breaklinks,colorlinks,citecolor=cvprblue]{hyperref}

\def\paperID{8041} 
\def\confName{CVPR}
\def\confYear{2025}

\newcommand{\ourmethod}{UA-Pose}

\title{{\ourmethod}: Uncertainty-Aware 6D Object Pose Estimation and Online Object Completion with Partial References}

\author{
Ming-Feng Li$^{1}$ \quad Xin Yang$^{4}$ \quad Fu-En Wang$^{4}$ \quad Hritam Basak$^{2}$ \\
\quad Yuyin Sun$^{4}$ \quad Shreekant Gayaka$^{4}$ \quad Min Sun$^{3,4}$ \quad Cheng-Hao Kuo$^{4}$ 
\vspace{15pt} \\
$^1$Carnegie Mellon University \quad $^2$Stony Brook University \quad $^3$National Tsing Hua University \\ \quad $^4$Amazon
}

\begin{document}
\maketitle
\begin{abstract}

6D object pose estimation has shown strong generalizability to novel objects. However, existing methods often require either a complete, well-reconstructed 3D model or numerous reference images that fully cover the object. Estimating 6D poses from partial references, which capture only fragments of an object's appearance and geometry, remains challenging.
To address this, we propose {\ourmethod}, an uncertainty-aware approach for 6D object pose estimation and online object completion specifically designed for partial references. We assume access to either (1) a limited set of RGBD images with known poses or (2) a single 2D image. For the first case, we initialize a partial object 3D model based on the provided images and poses, while for the second, we use image-to-3D techniques to generate an initial object 3D model.
Our method integrates uncertainty into the incomplete 3D model, distinguishing between seen and unseen regions. This uncertainty enables confidence assessment in pose estimation and guides an uncertainty-aware sampling strategy for online object completion, enhancing robustness in pose estimation accuracy and improving object completeness.
We evaluate our method on the YCB-Video, YCBInEOAT, and HO3D datasets, including RGBD sequences of YCB objects manipulated by robots and human hands. Experimental results demonstrate significant performance improvements over existing methods, particularly when object observations are incomplete or partially captured. Project page: \url{https://minfenli.github.io/UA-Pose/}

\end{abstract}

\section{Introduction}
\label{sec:intro}

6D object pose estimation, which determines the rigid 6-degree-of-freedom transformation between an object and a camera, is essential for various real-world applications such as robotic manipulation~\cite{wen2022catgrasp, wen2022you, zhuang2023instance,liu2023robotic} and augmented reality~\cite{marchand2015pose,su2019deep}. 
Recent research introduced \emph{model
-based} and \emph{model-free} approaches that can generalize to arbitrary novel objects. These methods push the boundaries of pose estimation for real-world applications and can be divided into two scenarios based on the object information available beforehand. \emph{Model-based} methods~\cite{shugurov2022osop,labbemegapose,zeropose,ausserlechner2023zs6d,nguyen2023gigapose,ornek2023foundpose,huang2024matchu,caraffa2024freeze,lin2024sam} require textured 3D CAD models of the object to estimate 6D object poses in 2D images by establishing 2D-3D correspondences. In contrast, \emph{model-free} methods leverage a large set of posed reference RGB images~\cite{liu2022gen6d}, RGBD images~\cite{he2022fs6d, wen2023foundationpose}, or video sequences~\cite{he2022oneposepp, sun2022onepose} to estimate object poses without the need for CAD models.


\begin{figure}[t]
\centering
\includegraphics[width=0.864\columnwidth]{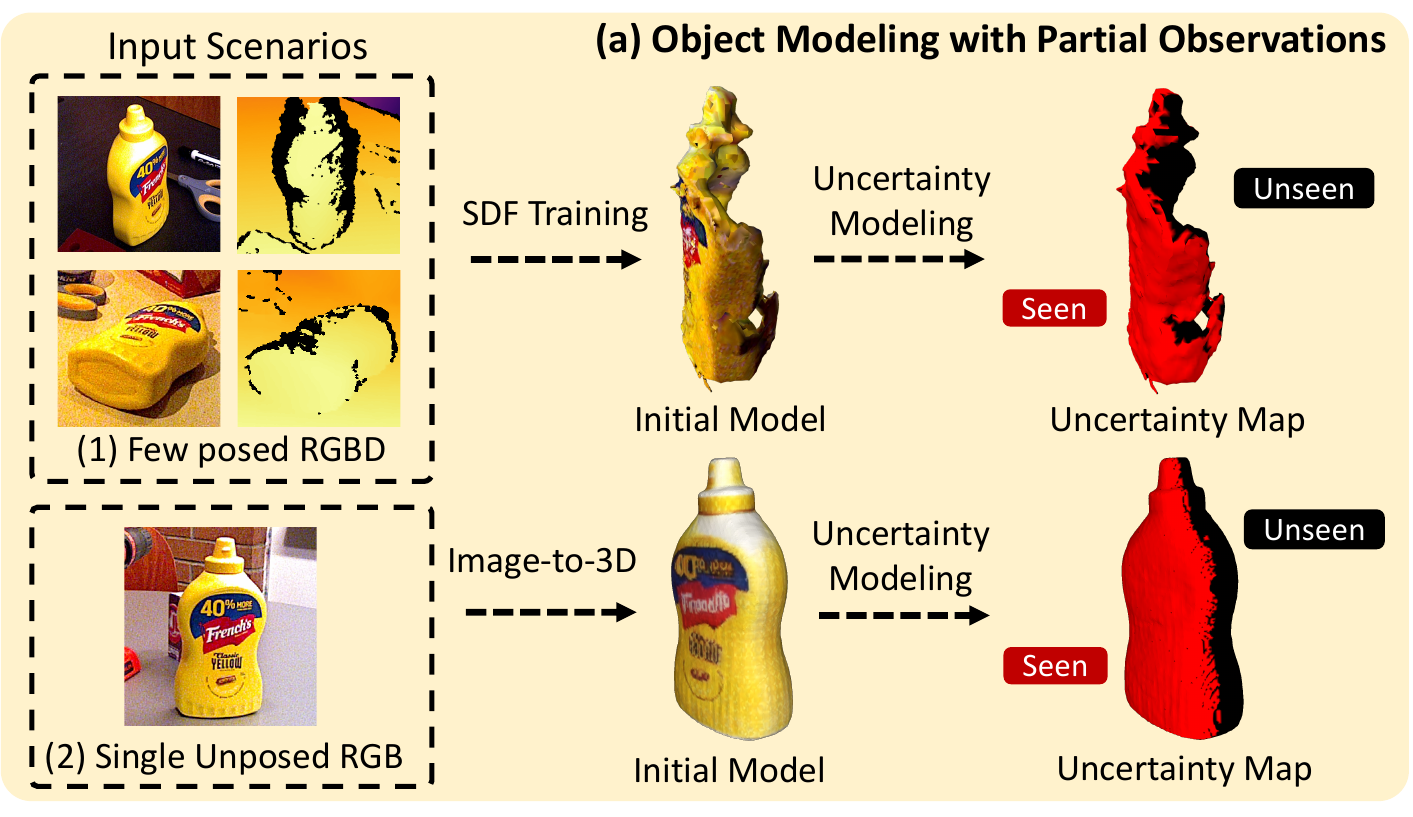}
\includegraphics[width=0.864\columnwidth]{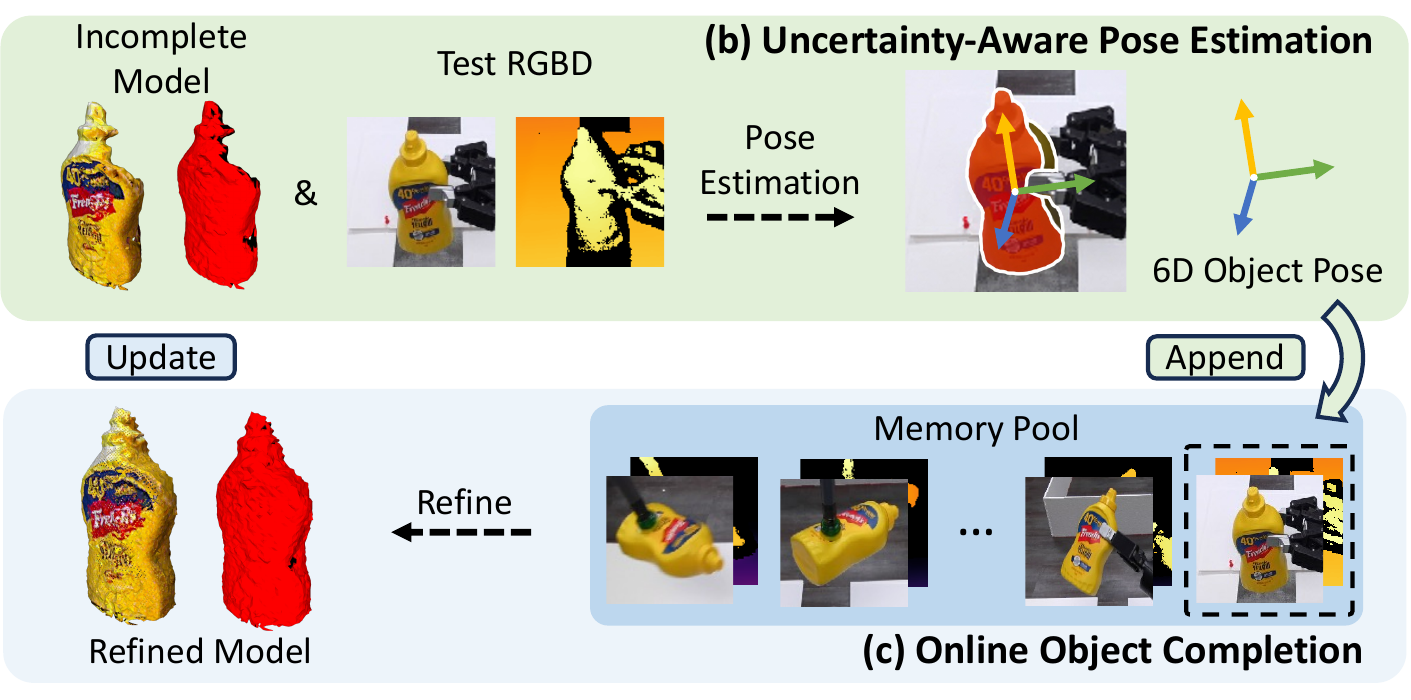}
\vspace{-0.4cm}
\caption{
    \textbf{Overview of {\ourmethod}.} 
    (a) We initialize an incomplete object 3D model $\mathcal{M}$ (\cref{sec:hybrid}), which is a hybrid mesh that combines texture, geometry, and uncertainty. 
    (b) Given a sequence of RGBD test images, our method estimates the 6D pose of the object in each test image (\cref{sec:ua-pose-estimation}) based on the incomplete model $\mathcal{M}$. (c) While more test images are captured, we store them with estimated poses in a memory pool, which will be used for online object completion (\cref{sec:object_completion}) to refine the incomplete model $\mathcal{M}$.
}
\vspace{-0.7cm}
\label{fig:teaser}
\end{figure}

To bridge \emph{model-based} and \emph{model-free} methods, FoundationPose~\cite{wen2023foundationpose} proposes a unified framework that supports both \emph{model-based} and \emph{model-free} setups, using 3D CAD models or posed RGBD images (referred to as references). While effective in generalizing to novel objects, FoundationPose and similar \emph{model-free} methods~\cite{he2022fs6d, he2022oneposepp, sun2022onepose} still depend on high-quality 3D models or sufficient posed RGBD images that capture the object from sufficient viewpoints. However, assuming the availability of high-quality 3D models is often impractical due to the vast number of objects that remain uncaptured or unmodeled in real-world environments. Likewise, obtaining adequate posed RGBD images is challenging, as real-world settings frequently involve limited viewpoints, occlusions from nearby objects, and complexities in estimating camera poses in dynamic scenes.


To address these challenges, we present {\ourmethod}, an uncertainty-aware approach for 6D pose estimation and online object completion as shown in~\cref{fig:teaser}. Our approach is designed to work with partial reference inputs that can include valuable meta-information for robotic tasks, such as grasping patterns, detection markers, or affordance labels. We consider two scenarios: (1) When a limited set of \textbf{reference RGBD images} with known poses is available, we initialize an incomplete 3D model using these images and poses; (2) When only a \textbf{single unposed RGB image} is available, we apply image-to-3D techniques to generate an initial object 3D model for pose estimation. By marking unseen regions as uncertain, we introduce a hybrid representation (\cref{sec:hybrid}) that integrates texture, geometry, and uncertainty. The uncertainty is utilized during our proposed uncertainty-aware pose estimation (\cref{sec:ua-pose-estimation}) to provide confidence for estimated poses. Additionally, we propose an uncertainty-aware sampling strategy to select informative images for online object completion, iteratively refining the object 3D model during testing (\cref{sec:object_completion}).

We evaluate our approach on the YCB-Video, YCBInEOAT, and HO3D datasets.
The results demonstrate significant improvements over existing methods and highlight the applicability of our uncertainty-aware approach in real-world scenarios where fully captured 3D models are often unavailable. 
To summarize, our contributions are:
\begin{itemize}
\item We propose an uncertainty-aware approach for 6D object pose estimation (\cref{sec:ua-pose-estimation}) and online object completion (\cref{sec:object_completion}), which improves both pose accuracy and object completeness when only partial references are available. 
\item We develop a hybrid representation (\cref{sec:hybrid}) incorporating uncertainty. The uncertainty enables confidence assessment in pose estimation and supports an uncertainty-aware sampling strategy for online object completion. 
\item We demonstrate pose estimation from a single RGB image by leveraging single-image-to-3D methods, which generate an initial model for pose estimation and provide augmented data for online object completion (\cref{sec:image-to-3d}). 
\item Our method supports different partial object references as input, which could be associated to meta-information, making it feasible for real-world applications.

\end{itemize}
\section{Related Work}
\label{sec:related}


\subsection{Model-based Object Pose Estimation}
Traditional \emph{model-based} pose estimation methods can be divided into two categories: \emph{instance-level}~\cite{he2021ffb6d,he2020pvn3d,park2019pix2pose,labbe2020cosypose} and \emph{category-level}~\cite{wang2019normalized,tian2020shape,chen2020learning,zhang2022ssp,lee2023tta,zheng2023hs}. Both types of methods assume that a textured CAD model of the specific object or similar object categories is available during training. However, these approaches are limited to objects seen during training, significantly restricting their applicability in real-world scenarios where many objects remain unseen or unmodeled.
To address this limitation, \emph{zero-shot} methods have been proposed~\cite{labbemegapose,shugurov2022osop,huang2024matchu,caraffa2024freeze,lin2024sam}, which estimate the pose of novel objects by providing their CAD models at test time. These methods use object CAD models to either render 2D images for feature matching and 2D-3D correspondence estimation or leverage point clouds from CAD models to estimate 3D-3D correspondences. While these approaches improve generalization to unseen objects, they still require access to high-quality 3D models at test time, which is often impractical in real-world settings.

\subsection{Model-free Object Pose estimation}
Model-free methods eliminate the need for an explicit textured CAD model. Gen6D~\cite{liu2022gen6d} introduces a pipeline to estimate object poses from RGB images with known poses. However, due to limited training data, Gen6D faces challenges with out-of-distribution test cases and images affected by occlusions. OnePose~\cite{sun2022onepose} and its extension OnePose++~\cite{he2022oneposepp} use structure-from-motion techniques to reconstruct 3D point clouds from sequential RGB images and employ pre-trained 2D-3D matching networks to estimate the pose of test images.
FS6D~\cite{he2022fs6d} employs a transformer-based architecture to extract features from multi-view RGBD images for pose estimation.
More recently, FoundationPose~\cite{wen2023foundationpose}, trained on a large synthetic dataset, presents a unified framework that supports both \emph{model-based} and \emph{model-free} setups. In its \emph{model-free} setting, FoundationPose requires posed reference images that capture the object from sufficient viewpoints to reconstruct 3D models for robust pose estimation. In general, these \emph{model-free} pose estimation methods rely on either sequential RGB images with known poses or densely posed RGBD images.
However, assuming pre-captured posed RGBD images for unseen objects is impractical for two key reasons: first, real-world applications often involve numerous unseen objects, making it infeasible to pre-capture and collect images for each object; and second, even if sufficient images of the novel object are available, accurately estimating camera poses in dynamic environments is challenging.



\subsection{Object Pose Tracking and Reconstruction}
Assuming test images for pose estimation are sequential, 6D object pose tracking methods~\cite{wen2021bundletrack, wen2023bundlesdf} leverage temporal cues to achieve efficient, smooth, and accurate pose predictions across consecutive frames by estimating relative pose transformations from each frame to the initial frame. For instance, BundleSDF~\cite{wen2023bundlesdf} is a 6D pose tracking method that estimates poses from sequential RGBD images while simultaneously training a signed distance field (SDF)~\cite{yariv2020multiview} representation to maintain consistent pose estimates over time. 
However, pose tracking methods are constrained to tracking object poses within sequential images from a single video, without support for using reference images for initial pose estimation.
In contrast, our method is designed to estimate object poses based on given object references, which may include metadata beneficial for robotic applications, such as grasping patterns, barcodes, or affordance labels. 


\subsection{Single-Image-to-3D and Pose Estimation}

Recently, single-image-to-3D methods have been developed to generate 3D models from a single image~\cite{xu2024instantmesh,long2023wonder3d,liu2023syncdreamer,shi2023zero123++,liu2023zero,qian2023magic123,voleti2025sv3d}. These methods offer potential for pose estimation when textured CAD models are unavailable. For instance, GigaPose~\cite{nguyen2023gigapose} explored using generated models for pose estimation; however, relying on generated 3D models that are not closely similar to real objects can lead to imprecise pose estimates. To address this, we apply a single-image-to-3D approach~\cite{xu2024instantmesh} to synthesize a generated model for early pose estimation. As real RGBD images are gradually collected during testing, they are used to create a refined object model, while the generated model from single-image-to-3D methods is leveraged to produce augmented data to improve object geometry in the uncaptured parts. Incorporating these augmented data into the SDF training yields more complete object models, ultimately improving the robustness and accuracy of pose estimation.

\section{Method}
\label{sec:method}

An overview of our method is illustrated in~\cref{fig:teaser}. We begin by initializing an incomplete object 3D model $\mathcal{M}$ using partial references, which can be either: (1) a few reference RGBD images with known poses or (2) a single unposed RGB image of the object. In this work, we propose a hybrid object representation (\cref{sec:hybrid}) that integrates texture, geometry, and uncertainty, serving as the incomplete model $\mathcal{M}$ for uncertainty-aware pose estimation (\cref{sec:ua-pose-estimation}).

Given a sequence of $k$ RGB test images, 
$\mathcal{I} = \{I_0, I_1, \dots, I_{k-1}\}$, corresponding depth maps 
, and an initial segmentation mask $m_0$ for the object of interest in the first image $I_0$, our method estimates the 6D pose of the object ${\xi}_i$ in each subsequent test image $I_i$ using the proposed uncertainty-aware pose estimation approach (\cref{sec:ua-pose-estimation}), based on the incomplete 3D model $\mathcal{M}$. During testing, newly captured test RGBD images and their estimated poses will be appended to a memory pool. These images will be used for online object completion (\cref{sec:object_completion}) to refine the incomplete object 3D model $\mathcal{M}$. 
Notably, when the input reference is a single unposed RGB image, we employ a single-image-to-3D approach~\cite{xu2024instantmesh} to initialize a generated model $\hat{\mathcal{M}}$ to assist in pose estimation and object completion. 
More details are described in~\cref{sec:image-to-3d}. 

\subsection{Background: Model-free Pose Estimation}


Given RGBD reference images of a novel object and their camera poses, the task of \emph{model-free} pose estimation is to determine the 6D object poses (i.e., the rigid 6D transformation from the object to the camera) of that novel object in test RGBD images. 
FoundationPose~\cite{wen2023foundationpose} is a foundational model for 6D pose estimation trained on a large-scale dataset. In the \emph{model-free} setup, it uses sufficient RGBD reference images (e.g., 16 RGBD images) that fully cover the object to train a Signed Distance Field (SDF) and then extracts a mesh by marching cubes as the 3D object model. Given the 3D object model to render object images, it introduces a two-stage pipeline with two specialized networks: a pose refinement network, which refines pose hypotheses by estimating the relative pose between the test image and rendered object images, and a pose selection network, which identifies the best pose by comparing the test image with rendered images based on these refined poses.
However, RGBD reference images that fully represent the object's appearance and geometry are usually unavailable in real-world scenarios. As a result, \emph{model-free} methods may struggle to provide accurate pose estimates when only incomplete object models or sparse observations are available.


\subsection{Hybrid Object Representation}
\label{sec:hybrid}


To address the limitations, we introduce a hybrid object representation that integrates the object’s texture, geometry, and uncertainty, as shown in~\cref{fig:sdf_training}. This uncertainty marks the seen and unseen regions of an incomplete 3D object model and provides confidence in pose estimations, helping to filter out less reliable results. 
Given a set of RGBD images, object masks, and their camera poses, we train an object-centric neural Signed Distance Field (SDF)~\cite{yariv2020multiview} to learn the partial object's appearance and geometry. 

We then extract a 3D mesh using the marching cubes algorithm to explicitly model uncertainty. Inspired by Visual Hull~\cite{laurentini1994visual}, we define object uncertainty based on visibility inferred from 2D object masks. By projecting these masks onto the mesh $\mathcal{M}$, we label regions of $\mathcal{M}$ seen in the reference images as “certain” and those unseen as “uncertain”, as shown in~\cref{fig:sdf_training}. This approach allows pose estimation to account for both the seen and unseen parts of the object.

\subsubsection{Neural SDF Training and Mesh Extraction} 
\label{sec:sdf_training}
\begin{figure}[t]
\centering
\includegraphics[width=0.99\columnwidth]{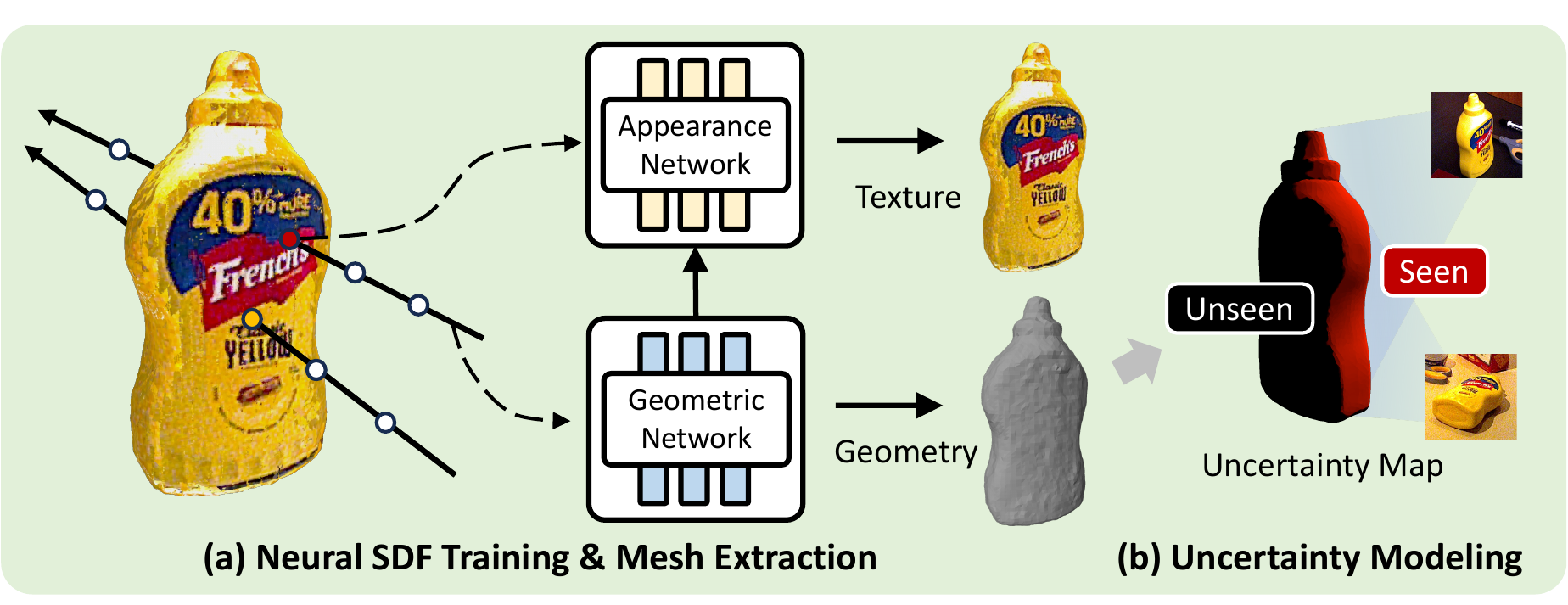}
\vspace{-0.3cm}
\caption{
    \textbf{Hybrid Object Representation Modeling.} We propose a hybrid object representation (\cref{sec:hybrid}) that integrates the object’s texture, geometry, and uncertainty. (a) First, a neural SDF is trained and extracted as a mesh representing the object’s appearance and geometry (\cref{sec:sdf_training}). (b) Then, we check the visibility of each mesh vertex from the viewpoint of each reference image to create the uncertainty map (\cref{sec:uncertainty_modeling}), which reflects the seen and unseen regions of an incomplete 3D object model.
}
\vspace{-0.5cm}
\label{fig:sdf_training}
\end{figure}

We follow SDF training methods similar to~\cite{wen2023foundationpose, wen2023bundlesdf} to train a neural SDF for object modeling. The SDF defines the object surface as the set of 3D points:
\begin{align} 
S = \{ x \in \mathbb{R}^3 \mid \Omega(x) = 0 \}. 
\end{align} 
where \(\Omega: \mathbb{R}^3 \rightarrow \mathbb{R}\) is the SDF geometry function that maps each 3D point \(x\) to a signed distance value \(s\) and \(s = 0\) indicates the object’s surface.
Given RGBD images, object masks, and their camera poses, we use a neural SDF to reconstruct the object’s appearance and geometry, represented by two networks~\cite{yariv2020multiview} as shown in Fig.~\ref{fig:sdf_training}. 
The geometric network $\Omega: x \mapsto s$ takes a 3D point $x \in \mathbb{R}^3$ as input and outputs a signed distance value $s \in \mathbb{R}$. 
The appearance network $\Phi: (f_{\Omega(x)}, d) \mapsto c$ takes an intermediate feature vector $f_{\Omega(x)}$ from the geometric network along with a view direction $d \in \mathbb{R}^3$ and outputs the color $c \in \mathbb{R}^3$ for the point.

\begin{figure*}[t]
\centering
\includegraphics[width=2.08\columnwidth]{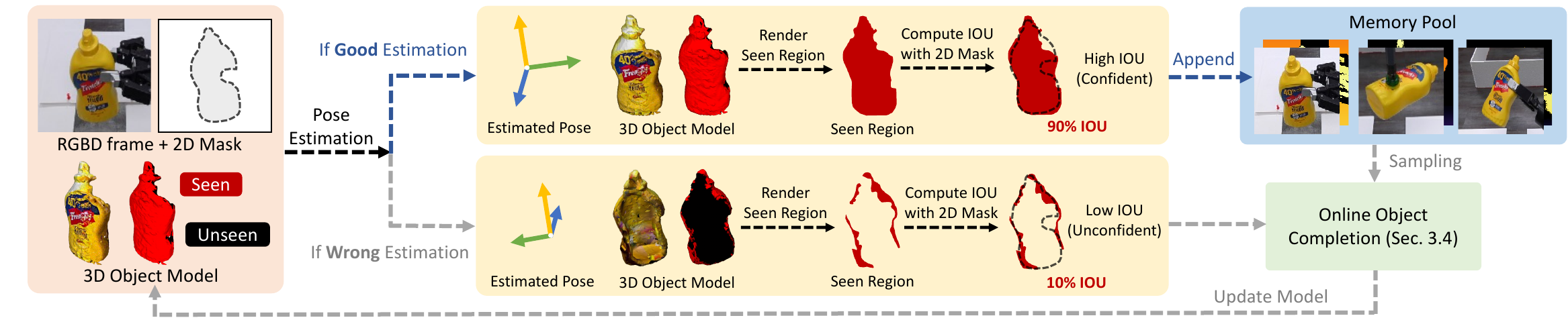}
\vspace{-0.25cm}
\caption{
    \textbf{Pipeline for Uncertainty-Aware Pose Estimation.} Through uncertainty modeling (\cref{sec:uncertainty_modeling}), we propose an uncertainty-aware pipeline to assess the confidence of estimated poses (\cref{sec:ua-pose-estimation}). For each estimated pose, we calculate its \emph{seen IoU}, which measures the overlap between the seen regions of the 3D model and the 2D object mask for the test image. When \emph{seen IoU} is high, the pose is deemed confident, and both the pose and its RGBD image are added to the memory pool for object completion. Conversely, if the estimated pose has a low \emph{seen IoU}, it is considered unreliable. In such cases, we employ an uncertainty-aware image sampling strategy to select images from the memory pool for online object completion (\cref{sec:object_completion}), thus enhancing pose estimation by refining the object model during testing.
}
\vspace{-0.45cm}
\label{fig:pipeline}
\end{figure*}


\noindent \textbf{Rendering.} 
Given the object pose $\xi$, we render an image by casting rays through each pixel. Along each ray $r$, 3D points are sampled at multiple positions as follows:
\begin{align} 
x_i(r) = o(r) + t_i d(r), \end{align}
\noindent where $o(r)$ denotes the ray origin and $d(r)$ denotes the ray direction from the camera to the object determined by the object pose $\xi$. The parameter $t_i \in \mathbb{R}_+$ determines the point position along the ray.
Following the approach in~\cite{wen2023bundlesdf}, the color $c$ of a ray $r$ is computed by integrating sampled points near the object surface through volumetric rendering:
\begin{equation} c(r) = \int_{z(r) - \lambda}^{z(r) + 0.5\lambda} w(x_i) \Phi(f_{\Omega(x_i)}, d(x_i)) dt, \label{eq
} \end{equation}
\begin{equation} w(x_i) = \frac{1}{1 + e^{-\alpha \Omega(x_i)}} \frac{1}{1 + e^{\alpha \Omega(x_i)}}, \end{equation}
\noindent where $w(x_i)$ is a bell-shaped probability density function~\cite{wang2021neus} based on the distance of each point from the implicit object surface and $\alpha$ adjusts the sharpness of the density distribution. $z(r)$ represents the ray’s depth from the depth image and $\lambda$ is the truncation distance. During volumetric rendering, $z(r)$ and $\lambda$ are used to exclude 3D points far from the object surface, enhancing rendering efficiency and preventing the SDF from overfitting to empty regions.

During training, we supervise this rendering by comparing it to the reference RGB images with the color loss:
\begin{align}
    \mathcal{L}_{c} = \frac{1}{|\mathcal{R}|} \sum_{r \in \mathcal{R}} \left\| c(r) - \bar{c}(r) \right\|_2,
\end{align}
where $\bar{c}(r)$ is the ground-truth color at the pixel through which the ray $r$ passes.
For geometry learning, we leverage depth information from the provided depth maps to supervise the neural (SDF) through two losses~\cite{wen2023bundlesdf}: the near-surface loss $\mathcal{L}_{\textit{s}}$ and the empty space loss $\mathcal{L}_{\textit{e}}$. $\mathcal{L}_{\textit{s}}$ encourage the SDF surface to align closely with depth values and $\mathcal{L}_{\textit{e}}$ enforces that points along the ray between the camera and the surface remain empty. Additionally, we apply an eikonal regularization loss $\mathcal{L}_{\textit{eik}}$~\cite{gropp2020implicit} to the near-surface SDF to enforce smoothness. The total loss is represented as:
\begin{equation}
\begin{aligned}
\mathcal{L} = w_{c}\mathcal{L}_{c} + w_{\textit{e}}\mathcal{L}_{\textit{e}} + w_{\textit{s}}\mathcal{L}_{\textit{s}} + w_{\textit{eik}}\mathcal{L}_{\textit{eik}},
\end{aligned}
\end{equation}
where $w_{c}$, $w_{e}$, $w_{s}$, and $w_{eik}$ are weighting factors.

To explicitly model uncertainty, we convert the implicit surface defined by the neural SDF into an explicit 3D mesh $\mathcal{E}=(V, C, F)$ with the vertices $V$, vertex colors $C$, and faces $F$.
To achieve this, we extract a mesh from the trained neural SDF representation using the marching cubes algorithm, which guarantees a closed-surface mesh including vertices from any viewing angle. This allows us to model the visibility of the object from any viewing angle even if some parts are not visible in the reference images.

Each vertex \(v_i \in V\) is assigned a color \(c_i \in C\), which is computed using the appearance function \(\Phi\), which maps the intermediate feature vector \(f_{\Omega(v_i)}\) from the geometry network and a view direction \(d \in \mathbb{R}^3\) to the RGB color: 
\begin{equation}
\begin{aligned}
c_i = \Phi(f_{\Omega(v_i)}, d).
\end{aligned}
\end{equation}

\subsubsection{Uncertainty Modeling}
\label{sec:uncertainty_modeling}

To model the uncertainty map that represents whether a vertex is visible from the viewpoints of input references, for each vertex \( v_i \in V \), we check its visibility from the viewpoint of each reference image \( \bar{I}_j \in \bar{\mathcal{I}} \) through the mesh rasterization process, which rasterizes 3D vertices onto 2D image pixels. We define an uncertainty score \( u(v_i) \) based on whether \( v_i \) is visible from any of the viewpoints in the reference images \( \bar{I}_j \).
This uncertainty score \( u(v_i) \in \{0, 1\} \) labels vertices on the mesh as ``certain" if they are seen in any reference image (\( u(v_i) = 0 \)) and as ``uncertain" if they remain unseen (\( u(v_i) = 1 \)). The resulting uncertainty map \(\mathcal{U} = \{ u(v_i) \mid v_i \in V \}\) is incorporated into our hybrid object representation \(\mathcal{M} = (\mathcal{E}, \mathcal{U})\), allowing the pose estimation process to distinguish between seen and unseen regions.

\noindent \textbf{Guidance from Uncertainty Map.} 
Given an object pose $\xi \in \mathbb{R}^{3 \times 4}$, the model $\mathcal{M}$ can used to render an RGB image $I^{\text{rend}}_{\xi} \in \mathbb{R}^{H \times W \times 3}$, a depth map $D^{\text{rend}}_{\xi} \in \mathbb{R}^{H \times W}$, a rendered object mask $m^{\text{rend}}_{\xi} \in \{0, 1\}^{H \times W}$, and an uncertainty map $U^{\text{rend}}_{\xi} \in \{0, 1\}^{H \times W}$ at a resolution of $H \times W$.
To evaluate the validity and confidence of an estimated pose, we introduce two metrics: \emph{uncertainty rate} and \emph{seen IoU}. The \emph{uncertainty rate} measures the proportion of the uncertain pixels over the total pixels within the rendered object mask:
\begin{equation}
\begin{aligned}
\frac{\sum (U^{\text{rend}}_{\xi} \odot m^{\text{rend}}_{\xi})}{\sum (m^{\text{rend}}_{\xi})},
\end{aligned}
\end{equation}
\noindent where $\odot$ denotes element-wise logical multiplication. 
The \emph{seen IoU} measures the overlap between seen (certain) regions of the rendered object mask and the object mask in the test image, which is defined as:
\begin{equation}
\begin{aligned}
\text{IoU}(\neg U^{\text{rend}}_{\xi} \odot 
 m^{\text{rend}}_{\xi}, m^{\text{test}}),
\end{aligned}
\end{equation}
\noindent where $m^{\text{test}}$ is the object mask from the test image, $\neg U^{\text{rend}}_{\xi} \odot m^{\text{rend}}_{\xi}$ identifies ``certain" pixels within the rendered object mask, and IoU represents the Intersection over Union function. 
Since pose estimation is more robust when based on seen regions of the 3D model, a high \emph{seen IoU} indicates high confidence in the pose estimation, whereas a low \emph{seen IoU} suggests that the estimation is not reliable.

\subsection{Uncentainty-aware Pose Estimation}
\label{sec:ua-pose-estimation}
After modeling the uncertainty, the hybrid object representation could be used in our uncertainty-aware pose estimation pipeline, as shown in~\cref{fig:pipeline}. In this work, we leverage the pose refinement and the pose selection modules of ~\cite{wen2023foundationpose} and assume that the test images are sequential. To obtain per-frame 2D object masks in test images for calculating the \emph{uncertainty rate} and \emph{seen IoU}, given first frame object mask $m_0$, the object masks of the following frames $\{m_1, m_2, \ldots, m_{k-1}\}$ is determined by off-the-shelf segmentation methods~\cite{yang2023track}.

\noindent \textbf{Pose Initialization.} Given an RGBD video, object masks for each frame, and an incomplete object model $\mathcal{M}$, we begin by estimating the object pose for the first frame by generating multiple pose hypotheses. For each hypothesis, the translation is initialized using the 3D point projected from the center of the 2D object mask, determined by the median depth value within this region. For rotation, we uniformly sample $N_v$ viewpoints from an icosphere centered on the object and orient the camera to face the object’s center. Subsequently, $N_{in}$ in-plane rotations are sampled and applied to each viewpoint, yielding $N_v \cdot N_{in}$ hypothesized object poses. Each object pose can be represented as $[\boldsymbol{R} \,|\, \boldsymbol{t}] \in \mathbb{SE}(3)$, where $\boldsymbol{R} \in \mathbb{SO}(3)$ is the rotation and $\boldsymbol{t} \in \mathbb{R}^{3}$ is the translation.

\noindent \textbf{Pose Refinement.} After generating $N_v \cdot N_{in}$ pose hypotheses, we apply the refinement module~\cite{wen2023foundationpose} to improve pose accuracy. Specifically, given a rendered RGBD image of the object from each pose hypothesis and the test RGBD image of the target object, the network outputs a pose update $\Delta \boldsymbol{R} \in \mathbb{SO}(3)$ and $\Delta \boldsymbol{t} \in \mathbb{R}^{3}$ to refine the pose hypothesis, aligning it more closely with the observed object pose in the test image. Each refined pose is represented as $[\boldsymbol{R}^{+} | \boldsymbol{t}^{+}] \in \mathbb{SE}(3)$, where $\boldsymbol{t}^{+} = \boldsymbol{t} + \Delta \boldsymbol{t}$ and $\boldsymbol{R}^{+} = \Delta \boldsymbol{R} \otimes \boldsymbol{R}$. This refinement process could be iteratively repeated. In this work, each pose is refined five times for pose estimation on the first frame and twice for subsequent frames.

\noindent \textbf{Pose Selection.}
Given a set of refined pose hypotheses, we apply our proposed metrics (\cref{sec:uncertainty_modeling}) to filter out unreliable poses: the \emph{uncertainty rate}, which measures the proportion of uncertain pixels within the rendered object mask, and the \emph{seen IoU}, which quantifies the overlap between previously seen (certain) regions of the object and the object mask in the test image. A pose hypothesis is discarded when its \emph{uncertainty rate} is above the threshold ${T}_{u}$ or its \emph{seen IoU} is below the threshold ${T}_{s}$, ensuring that only reliable hypotheses that adequately cover seen object regions are considered. At last, we use the pose selection module~\cite{wen2023foundationpose} to identify the most accurate pose as the final estimate.




\noindent \textbf{Pose Tracking.}
Given the sequential nature of RGBD video frames, object poses in adjacent frames are typically similar. For each frame following the initial one, we take the estimated pose from the previous frame as the sole pose hypothesis and refine it using the pose refinement module to generate the estimated pose for the current frame. This iterative process continues across the video sequence, leveraging temporal cues for stable and smooth pose estimation.

\noindent\textbf{Memory Pool.}
To enhance pose estimation by completing the object model during testing, we select test RGBD images with their estimated poses to incrementally refine the object representation. For efficiency, we introduce a memory pool $\mathcal{P} = \{{\xi}_0, {\xi}_1, {\xi}_2, \dots, {\xi}_{|\mathcal{P}|-1}\}$ that stores the most informative $|\mathcal{P}|$ viewpoints to maintain a compact yet diverse set of test image frames for object completion, where ${\xi}_i$ is the estimated pose corresponding to a test image $I_i$.
The first frame ${\xi}_i$ is automatically added to $\mathcal{P}$, establishing the canonical coordinate system for the novel object. Subsequent frames are added when their viewpoints contribute to the multi-view diversity in the pool. Specifically, before adding a new frame, we calculate the rotational geodesic distance between the new frame and the last frame added to the pool. A frame is included in $\mathcal{P}$ only if this geodesic distance exceeds a threshold $T_{geo}$, ensuring that each frame offers a unique viewpoint to enhance model completeness while keeping $\mathcal{P}$ compact.
Note that not all estimated poses are reliable, adding inaccurate results to the memory pool may cause noisy SDF training during object completion. To mitigate this, we introduce an image-filtering strategy to filter out low-confidence estimations. As illustrated in~\cref{fig:pipeline}, when a pose is estimated, its confidence is evaluated using our proposed metric, \emph{seen IoU}, which measures the overlap between the seen regions of the 3D model and the 2D object mask in the test image. Since a high \emph{seen IoU} indicates greater confidence in an estimated pose, frames are added to the memory pool only if their \emph{seen IoU} exceeds a threshold ${T}_{conf}$, ensuring that low-confidence poses are excluded.

\subsection{Uncentainty-aware Object Completion}
\label{sec:object_completion}

As shown in~\cref{fig:pipeline}, online object completion is triggered when the \emph{seen IoU} of the current test frame falls below a threshold ${T}_{complete}$, indicating that the current pose estimation may no longer be reliable.
As new frames are continuously added, the memory pool $\mathcal{P}$ can grow excessively large, which may reduce SDF training efficiency. To manage this, we perform uncertainty-aware sampling to select the $K$ most informative frames for object completion.

\noindent \textbf{Uncertainty-aware Sampling.} Early in the video sequence, when $|\mathcal{P}| \leq K$, all frames in the pool are included without selection. However, once the pool size exceeds $K$, we apply an uncertainty-aware image selection strategy to maximize the coverage of unseen object regions. This process selects a subset, $\mathcal{P}^{'} \subset \mathcal{P}$, that reveals the most unseen areas of the object.
The selection process begins by including the initial and most recent frames in the pool to establish the starting and ending viewpoints. Next, an iterative strategy is employed: at each iteration, we select the pose that reveals the largest unseen region. This continues until $K$ poses are selected.



\subsection{Leveraging Image-to-3D Generation}
\label{sec:image-to-3d}
\begin{figure}[t]
\centering
\includegraphics[width=0.99\columnwidth]{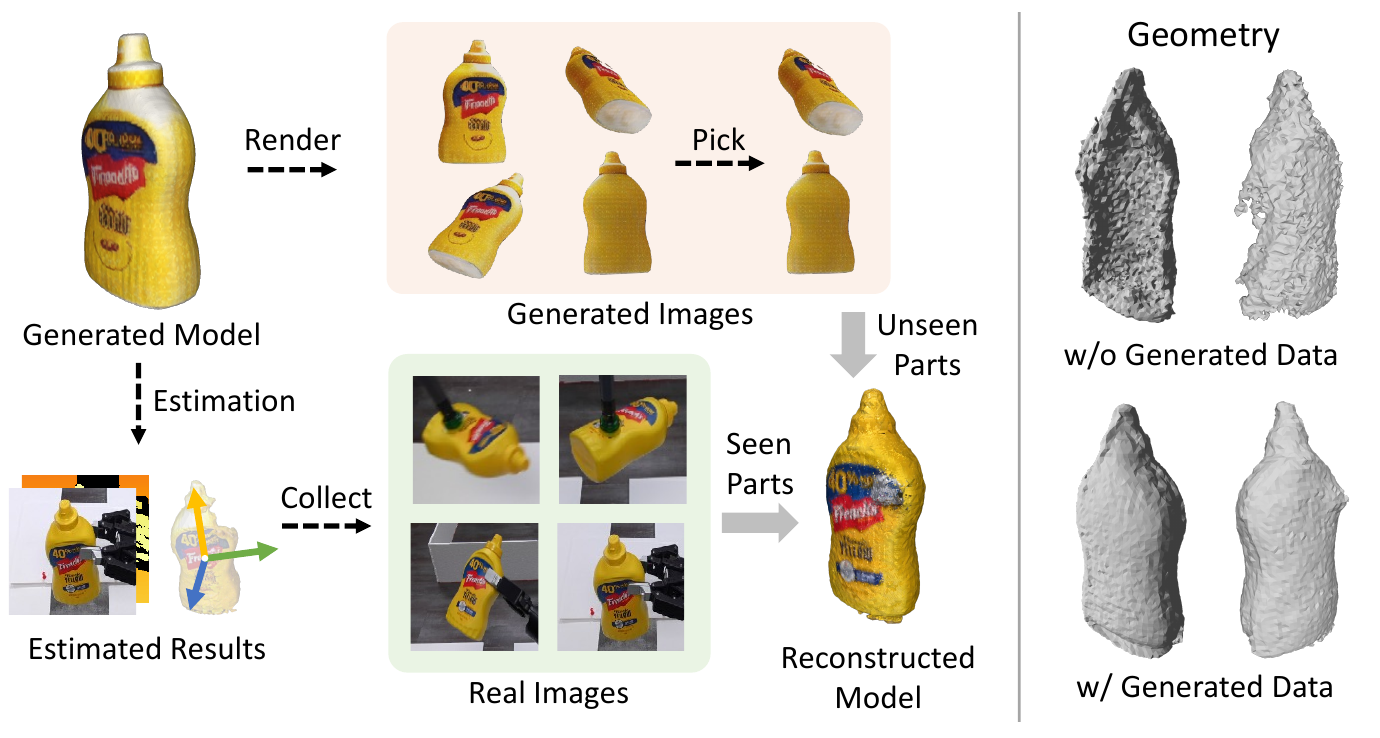}
\vspace{-0.5cm}
\caption{
    \textbf{Pipeline for leveraging generated models.} In~\cref{sec:image-to-3d}, the initial model generated by image-to-3D techniques is used to estimate poses for the first few frames. As more frames are captured, a refined object model is trained to replace the initial generated model, resulting in a more complete representation. To further leverage the generated 3D model, we render RGBD images from the image-to-3D generated model as augmented data for SDF training. This supervision enhances the object geometry in unseen regions, providing more complete information for pose estimation.
}
\vspace{-0.4cm}
\label{fig:image-to-3d}
\end{figure}

Given a single unposed RGB image, we employ image-to-3D techniques to generate an initial object 3D model $\hat{\mathcal{M}}$ for pose estimation. The viewpoint associated with the initial RGB image is labeled as “certain,” while other areas that are inferred by the image-to-3D method are marked “uncertain”.  The generated model is then rescaled to align with the object size in the test images. Specifically, we use the depth map and object mask from the first frame of the test image to approximate a rough model size. We then sample slightly larger and smaller models, generate pose hypotheses, and employ the pose selection module~\cite{wen2023foundationpose} to select the optimal pose hypothesis and corresponding model size.

Refer to~\cref{fig:image-to-3d}, the initial generated model is used to estimate poses for the first few test frames until the rotational geodesic distance between the initial pose and a new estimated pose exceeds a threshold $T_{gen}$. As more frames are captured, a refined object model is trained to replace the initial generated model, resulting in a more complete representation. To further leverage the generated 3D model, we render RGBD images from the image-to-3D generated model as augmented data for SDF training. This supervision enhances the object geometry in unseen regions, providing more complete information for pose estimation. See the supplementary materials for additional details.
\section{Experiment}
\label{sec:experiment}

\subsection{Datasets}
\noindent\textbf{YCB-Video~\cite{xiang2017posecnn}:} The YCB-Video dataset contains RGBD video sequences of everyday objects from the YCB objects, widely used for benchmarking 6D object pose estimation.

\noindent\textbf{YCBInEOAT~\cite{wen2020ycbineoat}:} YCBInEOAT extends the YCB-Video dataset with RGBD sequences of YCB objects manipulated by a dual-arm robot. This dataset focuses on real-world scenarios involving robot-object interactions.

\noindent\textbf{HO3D~\cite{hampali2020ho3d}:} The HO3D (Hand-Object 3D) dataset contains RGBD sequences of human hands interacting with objects, capturing complex hand-object interactions and occlusions.

\subsection{Experimental setup}
\label{sec:exp_setup}
\noindent\textbf{Baselines.} We evaluate our method against state-of-the-art RGBD \emph{model-free} 6D pose estimation methods~\cite{wen2023foundationpose, he2022fs6d, sun2021loftr} under two scenarios where partial object references are given as (1) A limited set of reference RGBD images (2 views) that include known poses. (2) A single unposed RGB object image. In scenario (2), we leverage image-to-3D techniques~\cite{xu2024instantmesh} to generate an initial object 3D model and iteratively estimate the object pose during testing.


\noindent \textbf{Metrics.}
We consider ADD and ADD-S metrics for 6D pose estimation, which measure the accuracy of estimated poses. The Area Under the Curve (AUC) is calculated and reported for both the ADD and ADD-S metrics, following the protocols in~\cite{xiang2017posecnn, wen2023foundationpose, wen2023bundlesdf}.
Furthermore, to demonstrate the effectiveness of our method for online object completion, we report the Chamfer Distance (CD), which measures the average distance between points of the reconstructed model and the ground-truth model.


\subsection{Comparison Results on YCB-Video}

We first evaluate {\ourmethod} on the YCB-Video~\cite{xiang2017posecnn} dataset, a widely used benchmark for 6D pose estimation, and compare our approach with state-of-the-art \emph{model-free} methods~\cite{sun2021loftr, wen2023foundationpose, he2022fs6d} with a limited set of reference images. 

As presented in~\cref{tab:ycbv}, 
our method significantly outperforms FoundationPose~\cite{wen2023foundationpose} with only 2 reference images in all metrics. Moreover, both our method and FoundationPose outperforms LoFTR~\cite{sun2021loftr} and FS6D-DPM~\cite{he2022fs6d} with 16 reference images with a large margin. This shows that FoundationPose is the strongest baseline for our further analysis. 
When only a single unposed RGB image of the object is available, we apply a single-image-to-3D approach~\cite{xu2024instantmesh} to generate an initial 3D model for pose estimation. Experimental results, shown in~\cref{tab:ycbv}, indicate that directly using the generated model with FoundationPose may not yield optimal results, as these models may not accurately represent real objects. In contrast, {\ourmethod} uses the generated model only for initial pose estimation and for rendering RGBD images as augmented data to aid object completion, as described in~\cref{sec:image-to-3d}. This approach reconstructs a model more closely aligned with the real object, leading to improved pose estimation accuracy.

\subsection{Comparison Results on YCBInEOAT}

\begin{table}[t]
    \centering
    \caption{\textbf{Quantitative results on the YCB-Video dataset.} “Comp.” indicates whether the method includes online completion, and “-” denotes that the method does not reconstruct object shapes. Results for~\cite{sun2021loftr, he2022fs6d} are adopted from~\cite{wen2023foundationpose}. Experiments for "single RGB + Image-to-3D" are conducted on a subset of the YCB-Video dataset. See supplementary materials for details.}
    \vspace{-0.3cm}
    \resizebox{0.48\textwidth}{!}{
    \begin{tabular}{c|c|c|ccc}

        \hline
        
        & \multirow{2}{*}{Input} & \multirow{2}{*}{Comp.} & \multicolumn{3}{c}{Mean} \\ 

        & & & ADD & ADD-S & CD \\
        \hline
        
        LoFTR~\cite{sun2021loftr} & 16 references & No & 26.2 & 52.5 & -\\ 
        FS6D-DPM~\cite{he2022fs6d} & 16 references & No & 42.1 & 88.4 & -\\

        FoundationPose~\cite{wen2023foundationpose} & 2 references & No & 87.4 & 94.3 & 0.57\\ 

        \ylwcell Ours & \ylwcell 2 references & \ylwcell Yes & \ylwcell \textbf{92.8} & \ylwcell \textbf{96.5} & \ylwcell \textbf{0.53} \\ 

        \hline
        \hline
        FoundationPose~\cite{wen2023foundationpose} & single RGB + Image-to-3D & No & 88.9 &	93.6 & 0.67 \\

        \ylwcell Ours & \ylwcell single RGB + Image-to-3D & \ylwcell Yes & \ylwcell \textbf{93.2} & \ylwcell \textbf{96.9} & \ylwcell \textbf{0.62} \\ 
        
        
        \hline 

    \end{tabular}
    }
\vspace{-0.2cm}
\label{tab:ycbv}
\end{table}

\begin{table}[t]
    \centering
    \caption{\textbf{Quantitative results on the YCBInEOAT dataset.} “Comp.” indicates that the method performs online completion. 
    }
    \vspace{-0.3cm}
    \resizebox{0.48\textwidth}{!}{
    \begin{tabular}{c|c|c|ccc}

        \hline
        
        & \multirow{2}{*}{Input} & \multirow{2}{*}{Comp.} & \multicolumn{3}{c}{Mean} \\ 

        & & & ADD & ADD-S & CD \\
        
        \hline

        FoundationPose~\cite{wen2023foundationpose} & 2 references & No & 68.52 & 84.80 & 0.60\\


        \ylwcell Ours & \ylwcell 2 references & \ylwcell Yes & \ylwcell \textbf{89.99} & \ylwcell \textbf{94.35} & \ylwcell \textbf{0.57} \\ 

        \hline
        FoundationPose~\cite{wen2023foundationpose} & single RGB + Image-to-3D & No & 83.39 & 90.66 & 0.61 \\
        \ylwcell Ours & \ylwcell single RGB + Image-to-3D & \ylwcell Yes & \ylwcell \textbf{89.24} & \ylwcell \textbf{93.99} & \ylwcell \textbf{0.60} \\ 

        \hline 
    \end{tabular}
    }
\vspace{-0.3cm}
\label{tab:ycbineoat}
\end{table}

We evaluated {\ourmethod} on the YCBInEOAT~\cite{wen2020ycbineoat} dataset to assess its performance in challenging robotic scenarios in real-world where objects are manipulated by robotic arms.

\begin{figure}[t]
\vspace{-0.35cm}
\centering
\includegraphics[width=0.99\columnwidth]{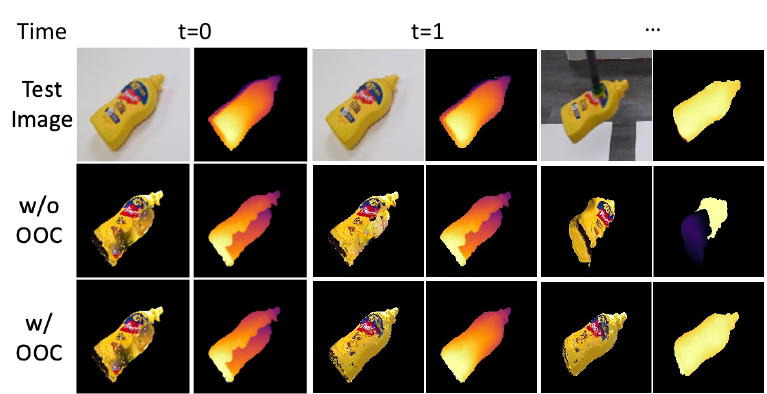}
\vspace{-0.25cm}
\caption{
    \textbf{Qualitative comparison on the YCBInEOAT.} 
    Object image (left) and depth map (right) pairs are rendered based on the estimated object poses using the incomplete model. Without uncertainty-aware mechanisms and Online Object Completion (referred to as “w/o OOC”), the incomplete model often leads to inaccurate estimations. In contrast, our method (referred to as “w/ OOC”) achieves robust pose estimation by incorporating uncertainty-aware mechanisms and online object completion.
}
\vspace{-0.4cm}
\label{fig:qualitative}
\end{figure}

\fong{In Table \ref{tab:ycbineoat}, when 2 reference images are provided, {\ourmethod} demonstrates superior performance compared to FoundationPose, highlighting the effectiveness of our uncertainty-aware mechanisms and online object completion for robust pose estimation in robotic scenarios. As illustrated in~\cref{fig:qualitative}, simply using the incomplete model generated from the partial references often leads
to inaccurate estimations. In contrast, our method achieves robust pose estimation with uncertainty-aware mechanisms and online object completion.}
\fong{When only a single unposed RGB image of the object is available, {\ourmethod} also consistently surpasses FoundationPose across all metrics. This demonstrates the flexibility of {\ourmethod} to utilize diverse partial references, which could carry useful meta-information.}

\subsection{Comparison Results on HO3D}

We evaluated {\ourmethod} on the HO3D dataset to assess its effectiveness in complex hand-object interaction scenarios where objects are frequently rotated and occluded.

As shown in~\cref{tab:ho3d}, when 2 reference images are provided, FoundationPose fails to make accurate estimations when objects in the test video exhibit frequent and strong rotations. In contrast, {\ourmethod} accurately estimates object poses under the challenging hand-object interaction scenarios.
As for utilizing the generated models from the single-image-to-3D approach for pose estimation, {\ourmethod} achieves more accurate estimations than FoundationPose. \fong{Note that even if the generated models in ~\cref{tab:ho3d} show lower Chamfer Distance (CD) than refined models obtained from object completion, directly using generated models for FoundationPose results in considerably lower ADD and ADD-S. This indicates that even though generated models have good geometry (low CD), geometry alone cannot accurately represent real objects for pose estimation because the textures of generated models are not good enough in unseen areas that are not covered by the initial partial references.}


\subsection{Ablation Study}

\noindent \textbf{Key Components.} \cref{tab:ablations} presents an ablation study for key components of {\ourmethod} on the YCBInEOAT dataset. All experiments start with the initialized model based on two reference images and are evaluated using ADD, ADD-S, and CD. Additionally, we report $N_{rebuild}$, the total number of times object completion is applied across test sequences to refine the incomplete model $\mathcal{M}$.
The “w/o object completion” setup disables online object completion, illustrating its importance for achieving accurate pose estimation. In “w/o uncertainty-aware object completion”, object completion is applied every time a new test image is added to the memory pool, instead of completing the model only when the \emph{seen IoU} threshold is below ${T}_{complete}$. This leads to poorer performance and a higher $N_{rebuild}$ count, increasing unnecessary computation for object completion. The “w/o image-filtering strategy” experiment omits the image-filtering strategy (\cref{sec:ua-pose-estimation}) for excluding unreliable poses, allowing low-confidence poses to be added to the memory pool and resulting in decreased pose accuracy.
In “w/o uncertainty-aware sampling”, we use the geodesic distance-based sampling strategy from~\cite{wen2023bundlesdf} instead of our proposed uncertainty-aware sampling strategy (\cref{sec:object_completion}) to sample images for object completion. The results demonstrate our proposed sampling method is better at selecting informative images for accurate pose estimation. Finally, the “Ours (full)” setup that incorporates all our key components achieves the best performance across ADD, ADD-S, CD, and the fewest $N_{rebuild}$ for object completion.

\begin{table}[t]
    \centering
    \caption{\textbf{Quantitative results on the HO3D dataset.} “Comp.” indicates that the method performs online completion.}
    \vspace{-0.3cm}
    \resizebox{0.48\textwidth}{!}{
    \begin{tabular}{c|c|c|ccc}

        \hline
        
        & \multirow{2}{*}{Input} & \multirow{2}{*}{Comp.} & \multicolumn{3}{c}{Mean} \\ 

        & & & ADD & ADD-S & CD \\

        \hline
        FoundationPose~\cite{wen2023foundationpose} & 2 references & No & 45.67 & 61.57 & 0.78 \\ 

        \ylwcell Ours & \ylwcell 2 references & \ylwcell Yes & \ylwcell \textbf{91.51} & \ylwcell \textbf{95.56} & \ylwcell \textbf{0.69} \\ 

        
        \hline
        FoundationPose~\cite{wen2023foundationpose} & single RGB + Image-to-3D & No & 72.06 & 87.87 & \textbf{0.76} \\

        \ylwcell Ours & \ylwcell single RGB + Image-to-3D & \ylwcell Yes & \ylwcell \textbf{83.23} & \ylwcell \textbf{93.59} & \ylwcell 0.88 \\

        \hline
                
    \end{tabular}
    }
\vspace{-0.2cm}
\label{tab:ho3d}
\end{table}

\begin{table}[t]
    \centering
    \caption{\textbf{Ablation study.} We demonstrate the importance of {\ourmethod}'s key components on the YCBInEOAT dataset.}
    \vspace{-0.3cm}
    \resizebox{0.45\textwidth}{!}{
    \begin{tabular}{c|ccc|c}

        \hline

        & \multicolumn{3}{c|}{Mean} & \multirow{2}{*}{$N_{rebuild}$} \\

        & ADD & ADDS & CD & \\ 
        
        \hline
        w/o object completion & 68.52 & 84.80 & 0.60 & - \\ 

        w/o uncertainty-aware object completion & 85.83 & 92.39 & 0.72 & 181 \\

        w/o image-filtering strategy & 87.70 & 93.37 & 0.71 & 87 \\ 

        w/o uncertainty-aware sampling & 87.88 & 93.39 & 0.68 & 59 \\ 

        \ylwcell Ours (full) & \ylwcell \textbf{89.99} & \ylwcell \textbf{94.35} & \ylwcell \textbf{0.57} & \ylwcell 58 \\
        
        \hline
                
    \end{tabular}
}
\vspace{-0.2cm}
\label{tab:ablations}
\end{table}

\begin{table}[t]
    \centering
    \caption{\textbf{Additional study.} We compare our method with FoundationPose~\cite{wen2023foundationpose} and pose tracking methods~\cite{wen2023bundlesdf, wen2021bundletrack} on the YCBInEOAT dataset. “Comp.” means the method performs online completion, and “*” denotes using cropped ground-truth meshes~\cite{wen2023bundlesdf}, which exclude invisible parts in the test sequence, to calculate CD.}
    \vspace{-0.3cm}
    \resizebox{0.45\textwidth}{!}{
    \begin{tabular}{c|c|c|ccc}

        \hline
        
        & \multirow{2}{*}{Input} & \multirow{2}{*}{Comp.} & \multicolumn{3}{c}{Mean} \\ 

        & & & ADD & ADD-S & CD \\
        
        \hline
 
        BundleTrack~\cite{wen2021bundletrack} & first-frame RGBD & No & 87.34 & 92.53 & 2.81* \\ 
        
        BundleSDF~\cite{wen2023bundlesdf} & first-frame RGBD & Yes & 86.95 & 93.77 & 1.16* \\ 

        FoundationPose~\cite{wen2023foundationpose}& first-frame RGBD & No & 74.34 & 84.58 & 1.12* \\ 

        \ylwcell Ours & \ylwcell first-frame RGBD & \ylwcell Yes & \ylwcell \textbf{88.38} & \ylwcell \textbf{93.82} & \ylwcell \textbf{0.75}* \\ 
        \hline 
    \end{tabular}
    }
\vspace{-0.45cm}
\label{tab:ablation_tracking}
\end{table}

\noindent \textbf{Additional Study.} 
\fong{When no external reference images are available, we demonstrate that we could initialize the object model for pose estimation using the first frame of the test RGBD image sequence. Besides the strongest \emph{model-free} pose estimation baseline~\cite{wen2023foundationpose}, we also compare our method with strong \emph{model-free} pose tracking approaches~\cite{wen2023bundlesdf, wen2021bundletrack, zhu2022nice} on the YCBInEOAT dataset. Refer to~\cref{tab:ablation_tracking}, our method achieves the highest performance across ADD, ADD-S, and CD metrics. Note that unlike pose-tracking methods which are limited to estimating relative poses from sequential frames within the same RGBD video, our method can incorporate object reference images as input for useful meta-information.}


\section{Conclusion}
We introduced {\ourmethod}, an uncertainty-aware approach for 6D object pose estimation that addresses the limitations of previous \emph{model-free} methods in the scenarios in which input object references are incomplete or partially captured. Incorporating mechanisms to handle uncertainty and leverage online object completion to complete object models, our method demonstrates substantial performance improvements on the YCB-Video, YCBInEOAT, and HO3D datasets. We demonstrate {\ourmethod} is a robust solution for practical real-world applications where full 3D object models and extensive image references are usually unavailable.


{
    \small
    \bibliographystyle{ieeenat_fullname}
    \bibliography{main}
}

\newpage
\section*{Supplementary Materials}

\setcounter{section}{0} 
\def\thesection{\Alph{section}}

\section{Implementation Details}  
 
The object images and uncertainty maps for pose estimation are rendered using the graphics pipeline~\cite{Laine2020diffrast} and Kornia~\cite{riba2020kornia}, and mesh rasterization for uncertainty modeling is performed with PyTorch3D~\cite{johnson2020accelerating}. For the pose refinement and selection modules described in Sec. 3.3 of the main paper, we utilize the publicly available checkpoints from~\cite{wen2023foundationpose}. Note that this version was not trained on diffusion-augmented data, which may lead to performance degradation compared to the expected results reported in~\cite{wen2023foundationpose}.
In this work, the neural Signed Distance Field (SDF) is trained using a simplified version of multi-resolution hash encoding~\cite{mueller2022instant}, leveraging the CUDA implementation from~\cite{wen2023bundlesdf}. The encoding is structured with 4 levels, each having feature vectors ranging from 16 to 128 in size and a feature dimension of 2. The hash table size is set to \(2^{22}\). Each training iteration processes a ray batch of 2048, with a truncation distance \(\lambda\) of 1 cm. Every SDF is trained for 500 steps, which completes within seconds.  
The geometry network \(\Omega\) is a two-layer MLP with a hidden size of 64 and ReLU activation for all layers except the last. It outputs a geometric feature \(f_{\Omega(\cdot)}\) with a size of 16. The appearance network \(\Phi\) is a three-layer MLP, also with a hidden size of 64. ReLU activations are used for all the layers except the last layer, where a sigmoid function maps the predicted colors to the range \([0, 1]\).

\subsection{Computation Time}
We evaluate the computation time of each module on a single NVIDIA RTX 3090 GPU. The full pose estimation procedure, which includes pose initialization, pose refinement, and pose selection, is completed within 2 seconds. Rendering uncertainty maps is highly efficient, requiring less than 0.01 seconds.
In practice, assuming sequential input test frames, the full pose estimation procedure is applied only to the first frame of the test image sequence. For subsequent frames, the estimated pose from the previous frame serves as the sole pose hypothesis, and only pose refinement is performed to update the pose for the new frame. This refinement process takes less than 0.02 seconds per frame, ensuring real-time performance.
When object completion is required, the full SDF training and Marching Cubes process is executed, taking approximately 30 seconds to generate the updated 3D model.

\subsection{Hyperparameters}
The hyperparameters and thresholds used in the main paper are set as follows.  
For pose initialization in both our method and the baseline implementation~\cite{wen2023foundationpose}, the sampled viewpoints (\(N_v\)) and in-plane rotations (\(N_{in}\)) are set to 42 and 12, respectively, yielding a total of 504 pose hypotheses during the hypothesis generation step. The number of pose refinement iterations is set to 5 for the first frame and 2 for subsequent frames.  
In the pose selection process, a pose hypothesis is discarded if its \emph{uncertainty rate} exceeds the threshold \({T}_{u}\) or if its \emph{seen IoU} falls below the threshold \({T}_{s}\), both set to 0.5.  
For storing frames and estimated poses for object completion, a frame is included in the memory pool \(\mathcal{P}\) (with a maximum of 30 frames) only if its geodesic distance exceeds the threshold \(T_{geo}\), set to 10 degrees. This ensures that each frame provides a unique viewpoint to enhance model completeness while keeping \(\mathcal{P}\) compact. Additionally, a new frame can be appended to the memory pool only if its \emph{seen IoU} exceeds the confidence threshold \(T_{conf}\), set to 0.5.  
Object completion is triggered when the \emph{seen IoU} falls below the threshold \(T_{complete}\), set to 0.7. This ensures that object completion is performed only when the model lacks sufficient completeness.  
For SDF training, the training weights are set as follows: \(w_c = 100\) for the color loss, \(w_e = 1\) for the empty space loss, and \(w_s = 1000\) for the surface loss.  


\subsection{Segmentation Methods}
In this work, the object mask for each test image is required to calculate \emph{seen IoU} and identify the region of the object for object completion. To obtain per-frame 2D object masks in test images, given the segmentation mask $m_0$ of the object of interest in the first-frame image $I_0$, the object masks of the following frames $\{m_1, m_2, \ldots, m_{k-1}\}$ are determined by off-the-shelf segmentation methods~\cite{yang2023track}.

\section{Datasets and Metrics}
\label{sec:intro}

\subsection{Preparing Partial References}
\noindent \textbf{Two RGBD references.} In our work, all experiments and baselines are conducted under the \emph{model-free} setting, where external RGBD images, along with their corresponding camera poses, are used as references for pose estimation. In~\cite{wen2023foundationpose}, 16 reference images per object are sampled from the YCB-Video training set~\cite{xiang2017posecnn} to form the subset $\mathbf{S}$, ensuring sufficient observations from diverse viewpoints. To ensure that the first frame of the test images is covered by the selected references for initial pose estimation, we manually select 2 reference images per object from $\mathbf{S}$ for the test sequences.

\noindent \textbf{Single unposed RGB reference.}
In scenarios utilizing single-image-to-3D methods~\cite{xu2024instantmesh} for pose estimation, we use a single unposed RGB image as input to generate an initial 3D object model. Specifically, we manually select one RGB image per object from the reference set $\mathbf{S}$ for each test sequence. This selected image serves as the sole external reference for generating the 3D model, without any additional depth or pose information. 

\noindent \textbf{Subset for the YCB-Video Dataset.}  
In this work, we leverage the pose refinement and selection models from~\cite{wen2023foundationpose} to estimate object poses using 3D object models generated by image-to-3D approaches~\cite{mueller2022instant}. 
However, we observed that current image-to-3D methods have difficulty generating accurate object models for certain objects. These poorly generated models result in failed pose estimations, introducing outliers into the average metrics. 
To ensure a fair evaluation, for the YCB-Video dataset~\cite{xiang2017posecnn}, we focus on a subset of the evaluation set for the ``single RGB + Image-to-3D" experiments, including commonly seen objects ``004\_sugar\_box,'' ``005\_tomato\_soup\_can,'' ``006\_mustard\_bottle,'' and ``019\_pitcher\_base.'' This selection ensures consistent and stable evaluation results across all baselines. For further discussion on the limitations of current image-to-3D methods and pose estimation models, please refer to~\cref{sec:limitations}.

\section{Leveraging Single-Image-to-3D Methods}
\label{sec: leverage-image-to-3d}
\noindent \textbf{Generate an object 3D mesh.}
To generate an initial 3D model of the object, we utilize InstantMesh~\cite{xu2024instantmesh}, a single-image-to-3D method that generates a coarse 3D mesh from a single RGB image. InstantMesh combines learned geometric priors with image features to infer the object's shape and produce a 3D mesh representation based solely on the input RGB image. Given a single unposed RGB reference, we employ InstantMesh to generate an initial 3D object mesh, denoted as $\hat{\mathcal{E}}^i_g=(V^i_g, C^i_g, F^i_g)$ where the vertices $V^i_g$, vertex colors $C^i_g$, and faces $F^i_g$. 

\noindent \textbf{Uncertainty map for the generated mesh.}
An uncertainty map, \(\hat{\mathcal{U}}^i_g\), is integrated into the mesh representation \(\hat{\mathcal{E}}^i_g\) to create a hybrid representation, \(\hat{\mathcal{M}}^i_g = (\hat{\mathcal{E}}^i_g, \hat{\mathcal{U}}^i_g)\), for uncertainty-aware pose estimation. The uncertainty map \(\hat{\mathcal{U}}^i_g\) is generated by marking the viewpoint corresponding to the initial RGB image as ``certain" and labeling areas inferred by the image-to-3D method as ``uncertain." This process involves projecting \(\hat{\mathcal{E}}^i_g\) onto the viewpoint of the reference image using mesh rasterization, which maps 3D vertices onto 2D image pixels. For each vertex, an uncertainty score \(u(v_i)\) is calculated based on its visibility in the reference image.
The uncertainty score \(u(v_i) \in \{0, 1\}\) is defined as follows: \(u(v_i) = 0\) if the vertex \(v_i\) is visible in the reference image, and \(u(v_i) = 1\) if the vertex is not visible. The resulting uncertainty map, \(\hat{\mathcal{U}}^i_g = \{ u(v_i) \mid v_i \in V^i_g \}\) is incorporated with $\hat{\mathcal{E}}^i_g$ to form the generated model \(\hat{\mathcal{M}}^i_g = (\hat{\mathcal{E}}^i_g, \hat{\mathcal{U}}^i_g)\).

\begin{figure}[t]
\centering
\includegraphics[width=\columnwidth]{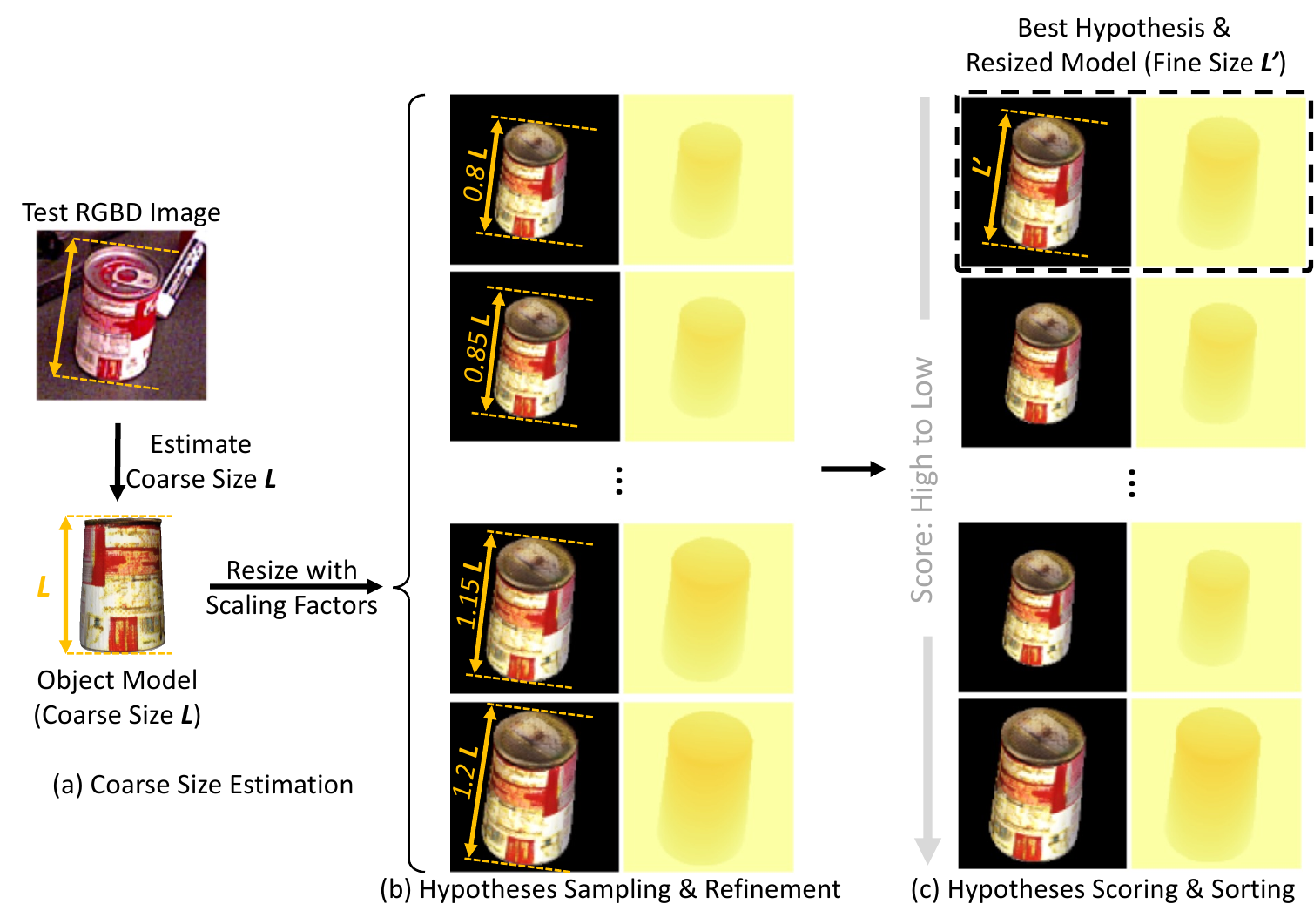}
\vspace{-0.5cm}
\caption{
\textbf{Pipeline for rescaling the generated model.} The generated object model must be rescaled to match the actual object size in the test images for accurate pose estimation.  
(a) In the first stage, the coarse size of the model is estimated using the first frame of the testing RGBD images. The maximum distance, \(L\), between the two farthest 3D points is computed to rescale the model to a rough size.  
(b) In the second stage, multiple scaling factors are uniformly sampled to slightly adjust the model size further, resulting in several resized models. For each resized model, pose hypotheses are generated and refined using the pose refinement model.  
(c) Finally, the selection strategy is applied to score the hypotheses and identify the optimal pose hypothesis and the fine size of the model.      
}
\vspace{-0.6cm}
\label{fig:mesh_resize}
\end{figure}

\begin{figure*}[t]
\centering
\includegraphics[width=2\columnwidth]{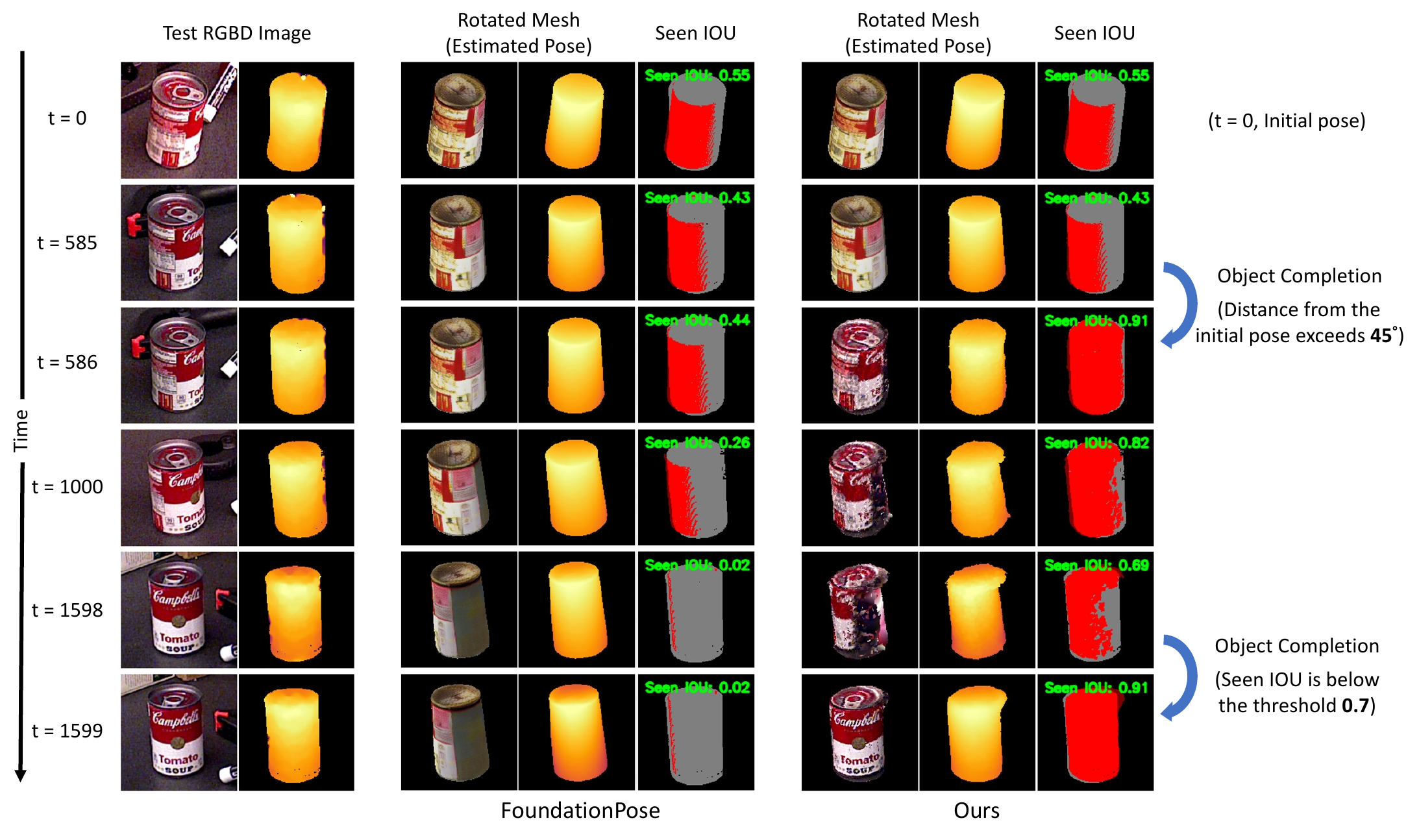}
\vspace{-0.4cm}
\caption{
\textbf{Qualitative results on the YCB-Video dataset.} We compare our method with FoundationPose~\cite{wen2023foundationpose} using an object 3D model generated from a single RGB image by the image-to-3D approach~\cite{xu2024instantmesh}. The columns, from left to right, display the test RGBD images (left: RGB, right: depth), results from FoundationPose, and results from our method. The object 3D models are visualized by rotating them based on the estimated poses. Additionally, the uncertainty map (red: seen region of the object 3D model; gray: object mask of the test image) and the \emph{Seen IoU} metric (indicating the overlap between the seen region of the object 3D model and the object mask) are shown.
In the final column, we demonstrate our iterative process of pose estimation and online object completion in our method, highlighting how an initial generated object 3D model is refined into a more complete and accurate object 3D model that better represents the real object's appearance and geometry. In contrast to directly using FoundationPose with the initial generated object 3D model (second column), which may not closely resemble the real object captured in the test images, our method maintains the completeness and correctness of the object 3D model while enhancing pose estimation accuracy. 
}
\vspace{-0.4cm}
\label{fig:qual_ycbv}
\end{figure*}


\noindent \textbf{Rescale the generated model.}
The generated model $\hat{\mathcal{M}}^i_g$ requires rescaling to match the object's actual size in the test images for accurate pose estimation.
To adjust the model to a size closer to the real object, we propose a two-stage coarse-to-fine process, as shown in the~\cref{fig:mesh_resize}. In the first stage, the coarse size is estimated using the depth map from the first frame of the testing RGBD images. Specifically, given the depth map and the object mask, 2D points within the object mask are projected into 3D space using the corresponding depth values. The maximum distance, $L$, between the two farthest 3D points is then calculated. The generated model (represented as a mesh) is scaled such that its diameter, which is defined as the distance between its two farthest vertices, matches this coarse object length $L$, resulting in the rescaled model $\hat{\mathcal{M}}^c_g$.
In the second stage, we refine the scaling to determine the fine object length, $L'$. Multiple scaling factors, $\mathcal{S} = \{0.8, ..., 1.2\}$ ($|\mathcal{S}|=11$), are uniformly sampled and multiplied by the coarse length $L$ to produce $|\mathcal{S}|$ resized models. For each resized model, $N'_v \cdot N'_{in}$ pose hypotheses are generated, where $N'_v=5$ represents the number of sampled viewpoints, and $N'_{in}=24$ denotes the number of sampled in-plane rotations. This process results in a total of $|\mathcal{S}| \cdot N'_v \cdot N'_{in}$ pose hypotheses.
We then apply the pose refinement and selection strategy described in Sec. 3.3 of the main paper to identify the optimal pose hypothesis. The corresponding resized model, $\hat{\mathcal{M}}^f_g$, is scaled to the fine object length $L' = s \cdot L$, where $s \in \mathcal{S}$. This refinement process is repeated iteratively three times, resulting in the final model $\hat{\mathcal{M'}}=(\hat{\mathcal{E'}},\hat{\mathcal{U'}})$ used for pose estimation.


\begin{figure*}[t]
\centering
\includegraphics[width=2\columnwidth]{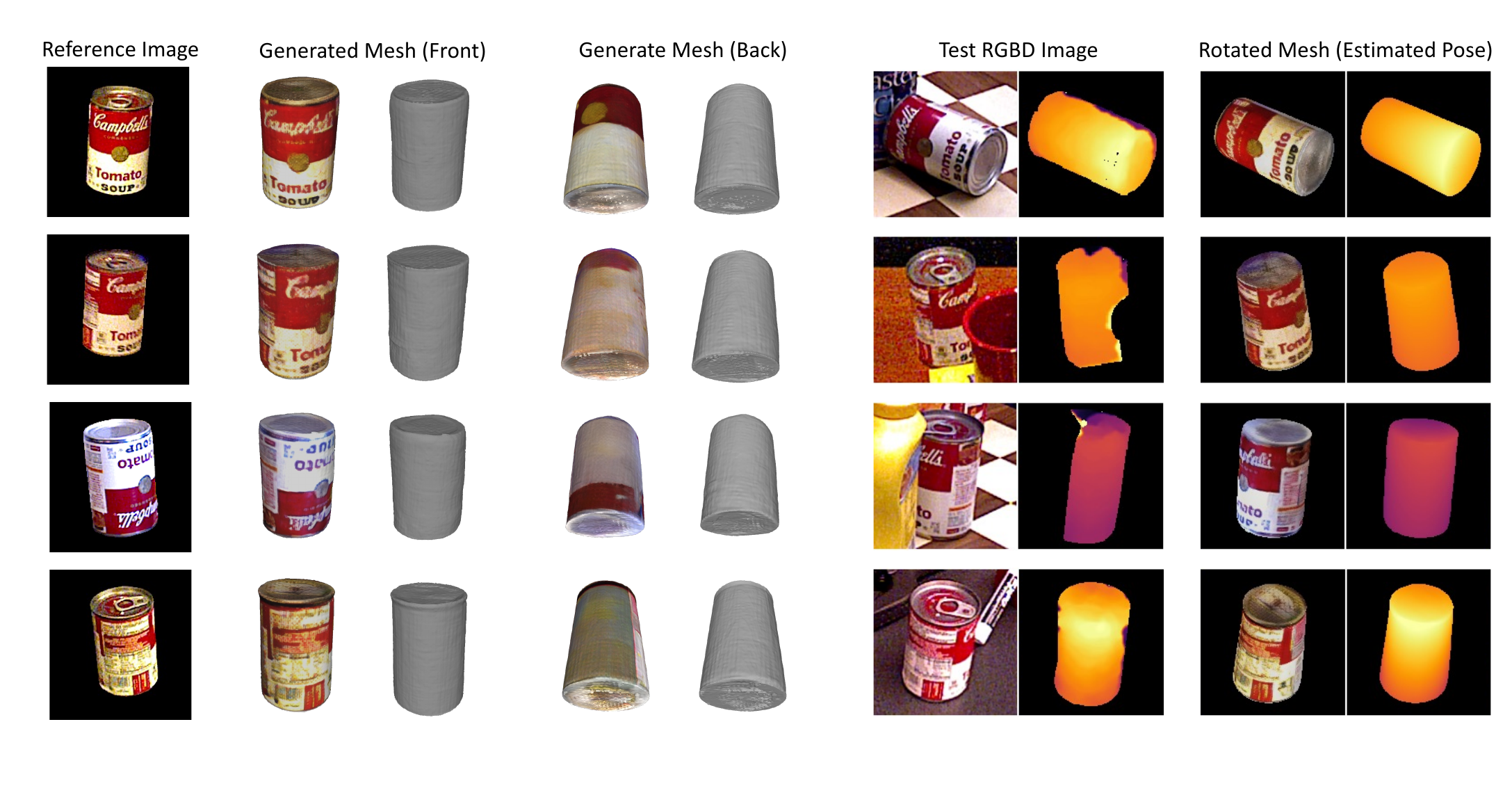}
\vspace{-0.85cm}
\caption{
    \textbf{Leveraging image-to-3D approaches for pose estimation across diverse reference images and test sequences.} We demonstrates the diversity and effectiveness of our method in utilizing the image-to-3D approach~\cite{xu2024instantmesh} to generate object 3D models for various test sequences. Each row represents a test sequence and includes the reference image, the generated object 3D model (front and back views), the test RGBD image, and the rotated mesh based on the estimated pose. The results highlight our method’s ability to successfully perform pose estimation across different test sequences, showcasing the potential of leveraging image-to-3D approaches for pose estimation using various single RGB images.
}
\vspace{-0.4cm}
\label{fig:diverse_generated_mesh}
\end{figure*}

\noindent \textbf{Generate augmented data for object completion.}
To further leverage the generated 3D model, we render RGBD images from the image-to-3D generated mesh \(\hat{\mathcal{E'}}\) as augmented data for SDF training. This supervision helps enhance the object geometry in unseen regions, providing more complete information for pose estimation. Specifically, we uniformly sample 24 viewpoints from an icosphere centered on the generated object mesh \(\hat{\mathcal{E'}}\), rendering synthesized object images \(\hat{\mathcal{I}} = \{\hat{I}_0, \hat{I}_1, \dots, \hat{I}_{23}\}\), along with their associated poses and 2D object masks, as augmented data for object completion.
During testing, the initial generated model \(\hat{\mathcal{M'}}\) is used to estimate poses for the first few test frames. When the rotational geodesic distance between the initial pose and a newly estimated pose exceeds a threshold \(T_{gen}\)(set to 45 degrees), a ``refined object model'' is trained to replace the initial generated model, resulting in a more complete and accurate representation. During this refinement, synthesized object images in \(\hat{\mathcal{I}}\) are used to enhance the geometry in unseen regions of the object. However, synthesized images may negatively impact SDF training if the corresponding regions have already been accurately captured by real images. Such overlap can introduce inconsistent supervision and degrade the final SDF quality.
To address this, we employ an uncertainty map \(\mathcal{U}\) to filter out unnecessary synthesized images. Before each SDF training iteration, we render the uncertainty map for the ``refined object model'' using the pose of each synthesized object image \(\hat{I}_i\) and compare the visible region with the 2D object mask of \(\hat{I}_i\). If more than 30\% of the pixels in \(\hat{I}_i\)'s 2D mask overlap with regions already marked as ``seen" in the ``refined object model'', the synthesized image \(\hat{I}_i\) is removed from \(\hat{\mathcal{I}}\). By filtering out redundant or conflicting synthesized data, this process ensures that synthesized images are used exclusively to supervise unseen regions.

\noindent \textbf{Difference with Gigapose.}
The prior work, GigaPose~\cite{nguyen2023gigapose}, investigated using object models generated by the single-image-to-3D method~\cite{long2023wonder3d} for pose estimation. In GigaPose, a 3D object model is created from the first frame of the ``RGBD" test images, with the depth map used to accurately scale the model. This generated 3D model is then directly applied for pose estimation across the entire test sequence. However, as discussed in Sec. 4 of the main paper, we demonstrate that such generated models often fail to represent real objects accurately, especially when objects rotate into unseen regions not covered by the initial partial references. Moreover, if the object is occluded in the first frame of the test image, GigaPose may struggle to generate a reasonably complete 3D model based on the cropped object image.
In contrast, we focus on a more challenging scenario where only a single external ``RGB" reference image is provided, without any depth information. ``RGB" images are more applicable to everyday use cases, such as using internet-sourced images as references. Moreover, these external reference images can carry valuable meta-information for robotic tasks, such as grasping patterns and detection markers, making them highly practical for real-world applications.

\section{Qualitative Results}

\subsection{Pose Estimation and Online Shape Completion}
In~\cref{fig:qual_ycbv}, we demonstrate the effectiveness of our method for pose estimation and online shape completion in the scenario where a single image is used as input and we leverage the image-to-3D approach~\cite{xu2024instantmesh} to generate the initial object 3D model for pose estimation. The experimental results show that directly using the generated model with FoundationPose~\cite{wen2023foundationpose} often fails to produce optimal results, as these models may not accurately represent real objects. In contrast, {\ourmethod} utilizes the generated model solely for initial pose estimation and for rendering RGBD images as augmented data to support object completion, as detailed in~\cref{sec: leverage-image-to-3d}. This iterative approach reconstructs an object 3D model that is more closely aligned with the real object.


\subsection{Diverse generated objects}

We showcases the ability of our method to utilize image-to-3D approaches for pose estimation with diverse reference images across different test sequences. As illustrated in~\cref{fig:diverse_generated_mesh}, each row corresponds to a unique test sequence. By leveraging image-to-3D techniques, our method achieves successful pose estimation with object 3D models generated based on single RGB images. This demonstrates the potential of leveraging image-to-3D approaches in enhancing pose estimation for real-world applications.


\section{Additional Analysis}
\noindent \textbf{Effects of the viewpoints of input reference images.} 
We conduct an experiment on the YCBInEOAT dataset~\cite{wen2020ycbineoat} to evaluate the effect of varying the viewpoints of input reference images, specifically in the scenario with two input reference images. For each test sequence, we select three distinct pairs of reference images captured from different viewpoints. The results show that FoundationPose~\cite{wen2023foundationpose} achieves average ADD-S, ADD, and CD scores of 75.61, 65.37, and 0.905 cm, respectively. In comparison, our method achieves 93.47, 85.26, and 0.691 cm, with an average of approximately 6 SDF reconstructions per test sequence. These results demonstrate that our approach is robust to variations in reference viewpoints and outperforms the baseline.

\begin{figure}[t]
\centering
\includegraphics[width=0.8\columnwidth]{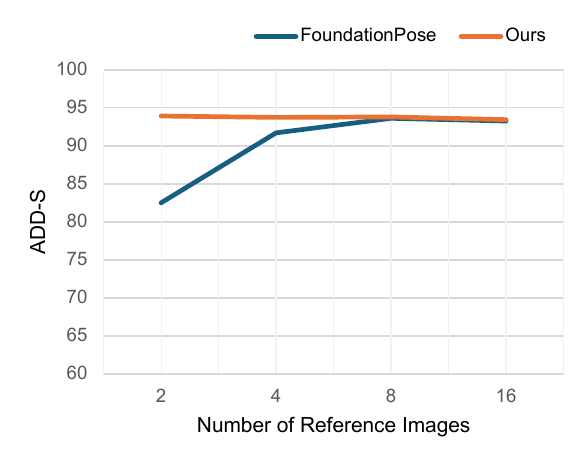}
\vspace{-0.5cm}
\caption{
\textbf{Effects of the number of reference images.} The ADD-S scores in the YCBInEOAT dataset with 2, 4, 8, and 16 reference images are reported. Our method demonstrates stable performance across different numbers of reference images, while FoundationPose~\cite{ornek2023foundpose} shows significant performance drops when fewer reference images (e.g., 2 and 4 views) are used.
}
\vspace{-0.6cm}
\label{fig:nviews}
\end{figure}

\noindent \textbf{Effects of the number of reference images.} 
To examine the effect of the number of reference views, we plot the ADD-S scores on the YCBInEOAT dataset using different numbers of reference images in~\cref{fig:nviews}. The ADD-S scores show that our method maintains consistent performance across different numbers of reference images, while FoundationPose~\cite{ornek2023foundpose} experiences significant performance drops when fewer reference images (e.g., 2 and 4 views) are used. Note that this experiment excludes the test sequence ``tomato\_soup\_can'' due to occasional pose estimation failures, as discussed in~\cref{sec:limitations}, which introduce outliers for comparison.


\begin{figure}[t]
\centering
\includegraphics[width=\columnwidth]{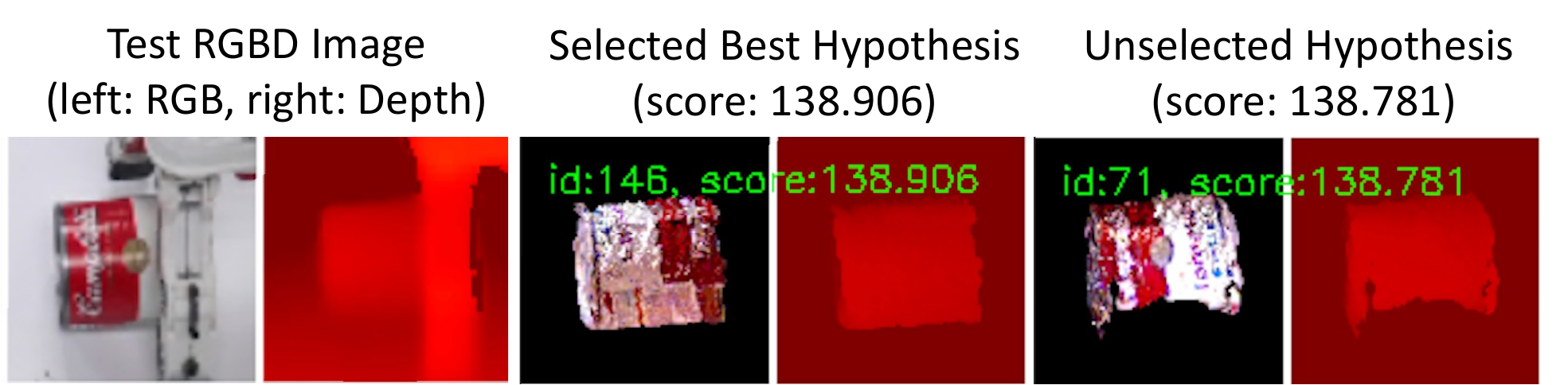}
\vspace{-0.5cm}
\caption{
\textbf{Incorrectly selected hypothesis.} Given a test RGBD image (left: RGB, right: depth), the pose selection module~\cite{wen2023foundationpose} scores all pose hypotheses and selects the one with the highest score as the pose estimation result. However, the selection module may occasionally assign a higher score to an obviously incorrect pose hypothesis (selected hypothesis, score: 138.906) while assigning a lower score to a more reasonable estimation (unselected hypothesis, score: 138.781). Such incorrect selections result in unstable pose estimation.
}
\label{fig:wrong_selection}
\end{figure}

\section{Limitations}
\label{sec:limitations}

\noindent \textbf{Foundation models for pose estimation.} 
As shown in~\cref{fig:wrong_selection}, the pose selection module of~\cite{wen2023foundationpose} may occasionally select an obviously incorrect pose hypothesis as the best. This issue likely arises because the pose selection model~\cite{wen2023foundationpose} was primarily trained on well-reconstructed object models that closely resemble real objects, rather than on incomplete, poorly reconstructed, or generated 3D models. To improve reliability, finetuning the pose selection model on synthesized data derived from incomplete object models could enhance its robustness and accuracy, reducing the occurrence of selecting clearly incorrect pose hypotheses.


\begin{figure}[t]
\centering
\includegraphics[width=0.8\columnwidth]{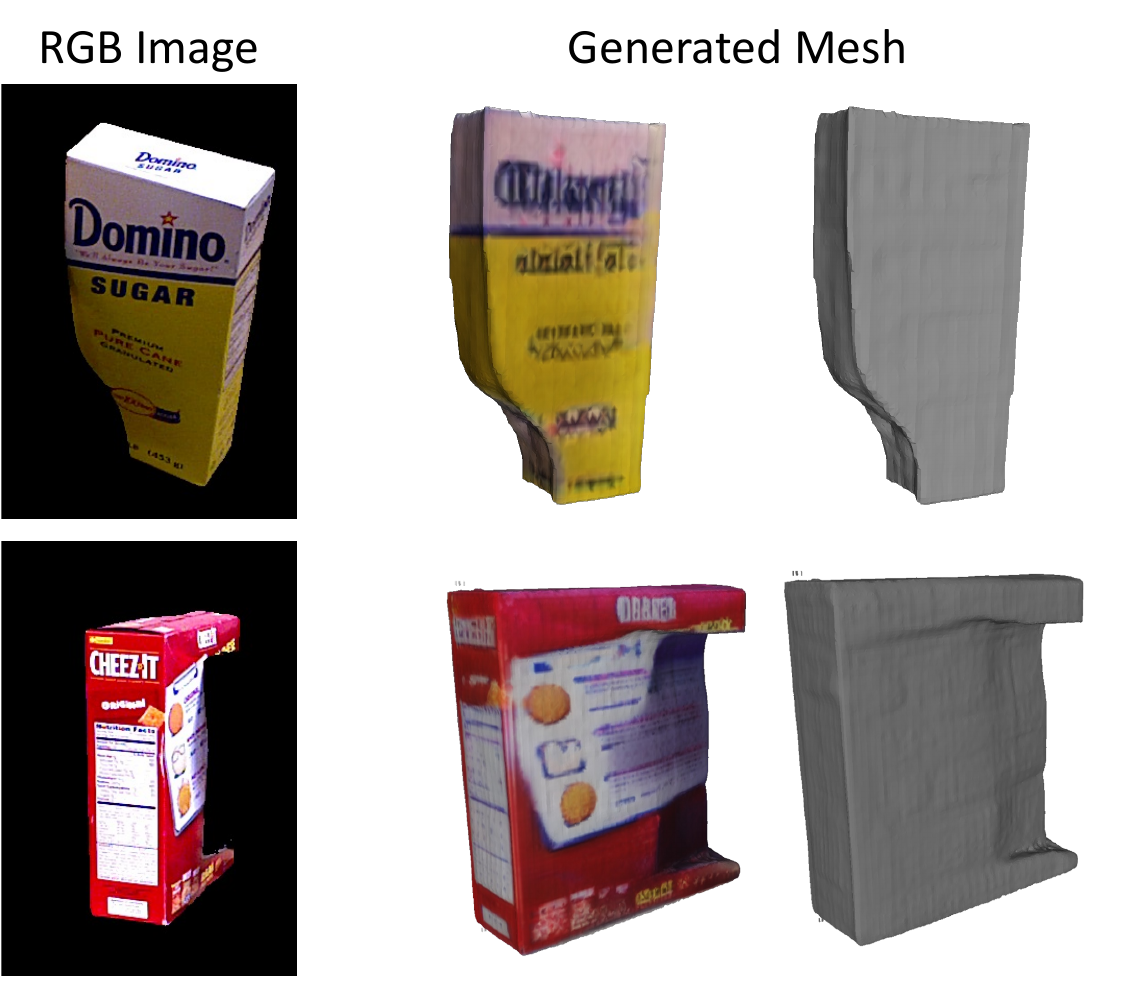}
\caption{
\textbf{Generated models from cropped object images.} We illustrate the limitations of single-image-to-3D approaches when relying on cropped or occluded reference images. The left column shows the RGB reference images and the right column displays the corresponding generated object 3D models. The results highlight that when the reference image is cropped or the object is partially occluded, the resulting 3D model is incomplete and fails to accurately represent the full object geometry.
}
\vspace{-0.4cm}
\label{fig:cropped_mesh}
\end{figure}

\noindent \textbf{Image-to-3D methods.}  
Image-to-3D approaches generally rely on high-quality object reference images. When the reference image is cropped or the object is partially occluded, the generated object 3D model is often incomplete, as shown in~\cref{fig:cropped_mesh}. Additionally, as highlighted in the experiments of the main paper, using a single unposed RGB image that captures an object from only a specific viewpoint often results in object 3D models that fail to accurately resemble real objects, particularly in their uncaptured regions.
Future work could address these limitations by integrating our pipeline with multi-view image-to-3D approaches to capture more comprehensive object information from different viewpoints. Alternatively, exploring image-to-3D methods that utilize RGBD inputs could better leverage depth information to infer more accurate and detailed object 3D geometry.

\end{document}